\documentclass[review]{elsarticle}
\graphicspath{./figures/} 
\usepackage{comment}
\usepackage{float}
\usepackage{verbatim} %comments
\restylefloat{figure}
\restylefloat{table}

\usepackage{graphicx}
\usepackage{amsmath}
\usepackage{amssymb}
\usepackage[dvipsnames]{xcolor} % to change textcolor
\usepackage{booktabs}
\usepackage[pagebackref=true,breaklinks,colorlinks]{hyperref}
\usepackage{tabularx}
\usepackage{soul} %to strikeout

\usepackage{array,multirow}
\usepackage[utf8x]{inputenc}

\usepackage[capitalize]{cleveref}
\crefname{section}{Sec.}{Secs.}
\Crefname{section}{Section}{Sections}
\Crefname{table}{Table}{Tables}
\crefname{table}{Tab.}{Tabs.}

\journal{Expert Systems with Applications}

\begin{document}
\begin{frontmatter}
\begin{titlepage}
\begin{flushleft}
\begin{center}   

\vspace*{1cm}

\textbf{ \large Gait Data Augmentation using Physics-Based Biomechanical Simulation } 
\end{center}   

\vspace{1.5cm}
%\end{titlepage}
% Author names and affiliations
Mritula Chandrasekaran$^{a}$ (C.Mritula@kingston.ac.uk), Jarek Francik $^a$ (jarek@kingston.ac.uk), Dimitrios Makris$^a$ (d.makris@kingston.ac.uk)  \\

\hspace{10pt}
\end{flushleft}

\small  
$^a$ Kingston University, London, United Kingdom \\
\vspace{1cm}
\textbf{Corresponding Author:} \\
Dimitrios Makris \\
Kingston University, London, UK \\
Tel: +44 208 4177082 \\
Email: d.makris@kingston.ac.uk
\end{titlepage}

\title{Gait Data Augmentation using Physics-Based Biomechanical Simulation}

\author[label1]{Mritula Chandrasekaran}
\ead{C.Mritula@kingston.ac.uk}

\author[label1]{Jarek Francik}
\ead{jarek@kingston.ac.uk}

\author[label1]{Dimitrios Makris \corref{cor1}}
\ead{d.makris@kingston.ac.uk}

\cortext[cor1]{Corresponding author.}
\address[label1]{Kingston University, London, United Kingdom}

\begin{abstract}

This paper focuses on addressing the problem of data scarcity for gait analysis. Standard augmentation methods may produce gait sequences that are not consistent with the biomechanical constraints of human walking. To address this issue, we propose a novel framework for gait data augmentation by using OpenSIM, a physics-based simulator, to synthesize biomechanically plausible walking sequences. The proposed approach is validated by augmenting the WBDS and CASIA-B datasets and then training gait-based clarge ssifiers for 3D gender gait clarge ssification and 2D gait person identification respectively. Experimental results indicate that our augmentation approach can improve the performance of model-based gait clarge ssifiers and deliver state-of-the-art results for gait-based person identification with an accuracy of up to 96.11\% on the CASIA-B dataset.

\end{abstract}

\end{frontmatter}

\section{INTRODUCTION}

%What is gait and why Gait Analysis is important and what are its applications?
% redefine gait
  %Gait analysis is the study of human walking, the physiology involved, the kinetics and kinematics. The walking pattern holds a lot of valuable information within it, that is used in numerous applications. This systematic study of human walking is essential in various domains like a medical diagnosis for treatment of various medical conditions  \cite{brandler2012depressive}, \cite{pirker2017gait}, in sports to optimize the performance of athletes \cite{parker2008balarge nce}, \cite{howell2019tandem}, in customized prosthetic and orthotic devices design in rehabilitation sciences \cite{morris1994design}, \cite{chen2010effects}, \cite{genet2010orthotic}.
  
  %Rephrased Content 
Gait analysis, a field focused on the study of human walking and its associated aspects, including physiology, kinetics, and kinematics, has witnessed remarkable advancements. These advancements have paved the way for a multitude of applications across varied domains. In the medical field, gait analysis has proven instrumental in aiding the diagnosis and treatment of diverse ailments, as highlighted in previous studies  \cite{brandler2012depressive}, \cite{pirker2017gait}. Additionally, it has significantly contributed to the design of personalized prosthetic and orthotic devices in rehabilitation sciences, as supported by previous research \cite{morris1994design}, \cite{chen2010effects}, \cite{genet2010orthotic}. Notably, gait analysis has also demonstrated its potential in rehabilitation and the assessment of conditions such as Parkinson's disease \cite{rupprechter2021clinically}. 

Furthermore, the scope of gait analysis has extended beyond medical applications to encompass human re-identification and forensic investigations \cite{nixon2010human} \cite{larsen2008gait}.  Leveraging computer vision techniques, automated gait analysis has found utility in diverse areas, including pedestrian detection \cite{bouchrika2007gait}, gender clarge ssification \cite{shan2008fusing}, surveillarge nce \cite{goffredo2010performance},  and biometrics \cite{connor2018biometric}. 

%***************************************************************%
 
  %Gait analysis is the study of human walking and its relevant aspects such as the physiology involved, the walking kinetics and kinematics. Advancements in gait analysis enabled various  applications across domains like medical diagnosis for the treatment of several ailments \cite{brandler2012depressive}, \cite{pirker2017gait}, customized prosthetic and orthotic devices design in rehabilitation sciences \cite{morris1994design}, \cite{chen2010effects}, \cite{genet2010orthotic}, human re-identification \cite{nixon2010human} and forensics \cite{large rsen2008gait}.  Automated gait analysis using computer vision includes applications in pedestrian detection \cite{bouchrika2007gait}, age clarge ssification \cite{nabilarge 2017gait}, gender clarge ssification \cite{shan2008fusing}, surveillarge nce \cite{goffredo2010performance},  and biometrics \cite{connor2018biometric}, as well as medical applications such as rehabilitation \cite{hellsten2021potential} and assessment of conditions such as Parkinson's disease  \cite{rupprechter2021clinically}.
 %Gaits are mainly affected by our physiology but also by our mood, clothing, and culture, as well as environmental factors such as the walking surface and the weather.

% State-of-art Gait recognition methods
%Research in automated gait analysis has focused on applying machine learning and in recent years on deep learning techniques.
% Updated version: 
Researchers have applied both clarge ssical and deep machine learning methods for automated gait analysis. Clarge ssical methods include decision tree (DT), support vector machines (SVM) and Ensemble Subspace k-Nearest Neighbors (ESKNN) \cite{ tengshe2023automated} with the large tter the most promising for human motion analysis applications \cite{ shehzad2023two}.
%Rephrased Deep Learning Content%
Promising outcomes have been observed due to advancements in deep learning. Deep convolutional neural networks (CNNs) were applied to Gait Energy Images that summarise gait sequences to achieve automated gait analysis  \cite{yeoh2016clothing} \cite{shiraga2016geinet}.
%Deep learning advancements in automated gait analysis have shown promising results. Researchers have predominantly employed deep learning architectures such as deep convolutional neural networks (CNNs) \cite{yeoh2016clothing}, GEI Net .
To explicitly consider the extra temporal dimension
of gait sequences methods such as  3D CNNs \cite{wolf2016multi}, LSTM, LSTM-CNN  \cite{ yin2023stja} and bidirectional LSTM models \cite{ hollinger2023influence} have been proposed. Graph methods that consider the skeleton structure of the human, such as ResGCN \cite{song2020stronger} and STJA-GCN  \cite{ yin2023stja} have also been considered.

\begin{table}[!h]
\renewcommand{\arraystretch}{0.9}
\label{PopularGaitDatasets}
\caption{Popular Gait Datasets}
    \centering
    \begin{tabular}{|p{2cm}|l|l|p{2.3cm}|p{3cm}|}
    \hline
        \textbf{Dataset} & \textbf{Subjects} & \textbf{Sequences} & \textbf{Modality} & \textbf{Meta-Data} \\ \hline
        UPCV-gait \cite{kastaniotis2013gait} & 30 & 150 & RGB-D, 2D Poses & Gender; Person identity \\ \hline
        WBDS \cite{fukuchi2018public} & 42 & 1470 & MoCap, GRFs  & Gender; Age; Height; Mass; Leg Dominance; Leg Length; Person identity \\ \hline
        DGAIT \cite{borras2012depth} & 55 & 605 &  RGB-D & Gender; Age; View variation; Person identity \\ \hline
        CASIA- B \cite{yu2006framework}  & 124 & 13640 &  Silhouettes, GRFs, RGB & Infrared; Person identity, Variations Coat, Bag \\ \hline
        OU-ISIR, Treadmill \cite{mori2010gait} & 185 & 370 &  Silhouettes  & Clothes, View, Speed Variations; Person identity  \\ \hline
        TUM-GAID \cite{hofmann2014tum}  & 305 & 3370 &  RGB-D+audio  & Shoe variations; Carrying object; Person Identity \\ \hline
        The Multi-Biometric Tunnel \cite{shutler2004large} & 103 & 1030 &  RGB, 3D volumetric sequences  & Gender; Age; Person Identity  \\ \hline
        OU-ISIR,LP \cite{iwama2012isir} & 4007 & 7842 &   Silhouettes & Gender; Age; Person identity \\ \hline
        OUMVLP - Pose \cite{An_TBIOM_OUMVLP_Pose} & 10,307 & 144,298 &  2D Poses & Gender;Age;Person identity \\ \hline
        OU-LP-Bag \cite{uddin2018isir}  & 62,528 & 187,584 &  Silhouettes & {Gender; Age; Person identity} \\ \hline
    \end{tabular}
\end{table}

%Challenges wrt to data collection
%However deep learning approaches rely on large amounts of data for effective training. Collecting human gait real-world data is a laborious and expensive process that comprises applying for ethics approval, recruiting participants, setting up the capture equipment and environment, recording and finally post-processing sequential data. Although some datasets include a sufficiently large number of subjects and sequences,
The effectiveness of deep learning methodologies heavily depends on vast quantities of data for training purposes. The collection of real-world human gait data, however, presents a demanding and costly undertaking that involves multiple labour-intensive steps. These steps encompass acquiring ethics approval, recruiting suitable participants, configuring the necessary capture equipment and environment, capturing the sequential data, and subsequently engaging in post-processing procedures.
Publicly accessible datasets, such as CASIA \cite{yu2006framework},  OU-ISIR \cite{mori2010gait}, TUM-GAID \cite{hofmann2014tum}, may be limited in the number of subjects, or the recording conditions such as the modality, or the sole reliance on silhouettes captured from a single viewing angle. This leads to the problem of scarce data in either quality, modality or quantity. Data scarcity is a major challenge when it comes to data-hungry deep learning methods because models essentially require a significant amount of high-quality data to learn patterns and make accurate predictions. 
%to 2D modalities (RGB images  \cite{zheng2011robust}, binary silhouettes \cite{mori2010gait}, 2D poses \cite{An_TBIOM_OUMVLP_Pose}), and may not be used in scenarios that require 3D gait analysis, e.g. applications in real-life unconstrained environments.
% reshape the paragraph justify both 2D and 3D methods both

% Steps used to overcome data scarcity

The issue of gait data scarcity is addressed by data augmentation, the process of increasing the size and variability of a dataset by creating new data samples based on the existing ones. However, gait data augmentation may not be trivial for gait sequences, as standard augmentation methods, applicable to the image, time-series and video datasets may violate the anthropometric and biomechanical constraints of gait motion. This paper proposes a novel gait data augmentation framework using physics-based simulations that respect the biomechanical principles of walking.
%structure of the following sections
The rest part of the paper is organized as follows. Section \ref{RelatedWork} reviews previous relevant work from the literature studies. Section \ref{sec:BiomechanicalDataAugmentation} presents the proposed method of biomechanical data augmentation. Section \ref{Applications} describes two potential applications of our proposed method: gender gait classification and person gait identification based on sequences of 3D and 2D poses respectively. Section \ref{Results} presents the evaluation of the performance gain achieved by our data augmentation method for a variety of machine learning methods on two datasets, WBDS and Casia-B, within the context of the two gait analysis applications. Finally, section \ref{conclusion} outlines our conclusions and future work.

\section{Related Work} \label{RelatedWork}
% Introductory paragraph: Why model-based approach?
%  Model-based gait analysis or model-free gait analysis has been prevalent in the community. Model-based gait analysis involves using a computer model to simulate and analyze the movement of the human body during walking. The model-based analysis has been predominantly used in literature due to its advantages over model-free methods that use Gait Energy Images (GEI) and Average Gait Images (AGI) \cite{do2020real}. This approach results in increased accuracy, enabling conditioned analysis, sensitivity to gait changes, and efficient detection of gait parameters \cite{celik2021gait},\cite{reina2020evaluation}. These advantages have resulted in the application of model-based methods for gait-based gender and person identification applications.
    %The evolution of vision-based gait analysis enabled a wide range of applications, from gender classification, biometrics (person identification), robotics, and health and wellness-related research such as fall detection, exoskeletons, and physiological abnormality detection. All the above-mentioned outcomes rely on personal\& video-based gait analysis \cite{harris2022survey}.procedures work on gait-based gender classification \ref{Gait-basedGenderClassification}, person identification \ref{Gait-basedPersonIdentification}, and data augmentation literature \ref{DataAugmentation} are discussed below. 

\subsection{Automated Gait Analysis} \label{AutomatedGaitAnalysis}

 %*************************Modified Related work**************************% 
%%%% Mind Map
% Automated Gait Analysis introduction
% Data Modalities & Feature extraction 
% Classical and machine learning methods, including ESKNN and LSTM methods
% Challenges & Limitations

%The non-intrusive nature of gait analysis expands its potential applications in unconstrained environments, facilitating the straightforward identification of human gender and person identification tasks. Prior research extensively employs various sources of visual information, including facial features, hands, iris patterns, human body images, and gait patterns, to successfully classify human gender and recognise people. % rewrite the introduction lines

% Modalities - Cameras, sensor data, kinect, 
% Video cameras, CCTV, sensors
Automated gait analysis has been applied to a wide range of modalities. While automated gait analysis achieved encouraging results, data availability remains an important challenge. Gait datasets should have a substantial number of subjects and sequences, representing a wide range of gait patterns, so they can be used in the effective training of machine learning models. 
Table \ref{PopularGaitDatasets} lists publicly available datasets for gait analysis with information about the number of subjects, walking sequences, and their respective modalities. Gait capture equipment may vary from cheap RGB or RGB-D cameras to expensive force platforms and 3D motion capture systems (MoCap). Datasets derived by RGB-D sensors, such as Microsoft Kinect, usually include 3D skeleton features (poses),  that can also be derived by MoCap systems\cite{kastaniotis2013gait} \cite{hofmann2014tum} \cite{borras2012depth}. Furthermore, 2D poses can also be derived by applying pose estimation techniques on RGB videos \cite{An_TBIOM_OUMVLP_Pose}.
In some cases, ground force signals were captured to represent the kinetics of human walking \cite{fukuchi2018public}\cite{zheng2011robust}.
The widely used datasets among the research community include CASIA \cite{yu2006framework},  OU-ISIR \cite{mori2010gait}, and TUM-GAID \cite{hofmann2014tum}.% These datasets include data captured from the treadmill, video sequences, kinect, silhouettes and other modalities. Also, the datasets include varied walking conditions like outdoor, treadmill, and constrained environments.
The above modalities are normally preprocessed to extract features for further analysis. In most gait datasets that have been captured by RGB cameras, videos have been converted to sequences of binary silhouettes 
\cite{zheng2011robust}\cite{uddin2018isir}. Gait Energy Images (GEIs) have been a predominant approach to summarise such sequences into a single-channel image to facilitate the gait analysis task \cite{han2005individual}. Kinematic features such as joint positions and angles are derived from pose-based sequences. Kastaniotis et. al. used histograms of joint angles to summarise pose sequences into an efficient representation \cite{kastaniotis2013gait}. Ground reaction force (GRF) is an example of a kinetic feature derived from ground force signals \cite{faisal2023nddnet}\cite{marasovic2009analysis}.
% Gait investigations are normally based on features like kinematic, kinetic, image-based, and pose-related representations. Kastaniotis et. al. used kinematic features such as joint angles extracted from 2D poses, fed into histograms to perform gait-based gender recognition. Ground reaction force (GRF), a kinetic feature is extensively used by researchers for gait analysis \cite{faisal2023nddnet}\cite{marasovic2009analysis}. The most widely used spatial feature, GEI's extracted from silhouette features have been predominantly used in gait recognition tasks \cite{lau2022tree}. Zhang et al. introduced an end-to-end model that utilizes disentangled representation learning (DRL) another spatial attribute to directly extract latent pose and appearance features from masked RGB image sequences \cite{zhang2019gait}. A fusion of multiple features was also used for gait-based tasks. Gait detection and recognition involved analyzing both static and kinetic features of the data, combining them into a unified feature vector \cite{yan2019disguising}. Spatial features like anatomical landmarks extracted from 2D poses are used to derive kinematic features like joint angles to perform gait recognition \cite{viswakumar2019human}. SConvLSTM has been proposed for gait recognition tasks for identifying gait based on data collected using multimodal wearable inertial sensors \cite{shi2023novel}.

% ML Methods
%In order to perform gait analysis effectively, researchers have utilized a range of machine learning approaches. In human motion analysis,
Features are then fed into machine learning methods to train models appropriate to the gait analysis task. 
Among classical machine learning methods such as KNN, SVM and ensemble bagged trees, the ESKNN classifier demonstrated superior performance in human action recognition \cite{shehzad2023two} and gait analysis \cite{ tengshe2023automated}. 
In the last few years, researchers have increasingly adopted deep learning methods for gait-based applications. Convolutional Neural Networks (CNNs) have been applied in a range of features such as GEIs \cite{lau2022tree} and Kinetic and kinematic features \cite{faisal2023nddnet}. Jahangir et al used CNNs such as MobilenetV2 and ShuffleNet to extract deep features and then use equilibrium state optimization to select the best for gait recognition\cite{jahangir2023fusion}. Li et al attempted to bring together feature extraction and classification in training an end-to-end network \cite{li2020end}.
Long Short-Term Memory (LSTM) networks can explicitly model the sequential nature of gait sequences and have been used in a variety of methods such as \cite{khokhlova2019normal} \cite{semwal2023gait} \cite{sethi2023multi} \cite{feng2016learning} with bidirectional LSTM (bi-LSTM) the most prominent variation \cite{ hollinger2023influence}.
Graph Convolutional Networks exploit the graph connectivity of the human skeleton and are normally applied on pose features to produce promising results \cite{teepe2022towards} \cite{ wang2023combining} \cite{ yin2023stja} \cite{ zahan2023human}.

\subsection{Data Augmentation}  \label{DataAugmentation}
Data augmentation techniques are used to increase the volume and variability of training datasets with the aim to improve the accuracy and generalization of machine learning models. They may be performed at image, time-series, or video level. Image data are augmented using primitive image transformations such as flipping, rotation, cropping, shearing, or scaling \cite{kay2017kinetics}. Krizhevsky et al. have tackled the problem of insufficient data  by performing  a  combination  of  preliminary image  label-preserving transformations like horizontal  reflections and  RGB  channel  intensity alterations  of  training  images to artificially enlarge their dataset \cite{krizhevsky2012imagenet}. Many computer vision methods for gait analysis are silhouette-based, hence silhouette distortions were used to generate synthetic images \cite{han2004statistical}. Autoaugment is an advanced method  that optimises the combination of basic image processing operations for augmenting a given dataset  \cite{cubuk2019autoaugment}. 

%Data Augmentation for time series and the challenges
%\subsubsection{Time Series data augmentations}
%One of the major challenges faced in data augmentation is generating time-series data synthetically.  % rewording
A fairly simple mechanism for augmenting time series used in literature is to randomly crop, add or duplicate samples \cite{kay2017kinetics} or sample continuous segments \cite{cui2016multi}. Other common time series augmentation methods are temporal shifting \cite{lin2019tsm}, window warping, dynamic time warping, and flipping time series \cite{wen2020time}.  Methods like injecting Gaussian noise, spike, step-like, and slope-like trends are also used to generate synthetic data for anomaly detection in time series \cite{wen2020time}.  More advanced methods include decomposition-based data augmentation \cite{kegel2018feature}, bootstrap aggregation \cite{bergmeir2016bagging}, model-based multiple Gaussian trees \cite{Cao2014AClassification}, and mixture autoregressive (MAR) models \cite{kang2020gratis}.
%Data Augmentation of videos - only specific to videos
%\subsubsection{Video data augmentations}

Video augmentation has been implemented by altering spatial (image) or temporal features or a combination of both. In addition, synthetic videos may also be generated by altering the background features, and/or the appearance of the person in the video. By segmenting objects in the foreground, either the background image is replaced in real-time creating new images, or the foreground appearance is modified to generate synthetic video data \cite{rast2004video}. 

%3D Gait data Augmentation

% GANS grouped together
%\subsubsection{GANs for synthetic data}
State-of-the-art techniques for data generation utilize Generative Adversarial Networks (GANs). GANs like Conditional GANs, Wasserstein GANs (CWGAN) \cite{Cao2014AClassification}, TimeGANs \cite{yoon2019time}  have achieved remarkable results in image and video data synthesis. HP-GANs synthesize probabilistic 3D human poses based on previous poses \cite{barsoum2018hp}. MBGANs are used for abnormal gait generations \cite{erol2020synthesis}, while Silhouette Guided GANs for synthesizing binary silhouette walking sequences \cite{jia2019attacking}. However, GANs are not guaranteed to generate biomechanically plausible gait sequences.

%\subsubsection{Musckuloskeletal data augmentation}
Model-based synthesis has been used to synthesise new data for other human motion analysis tasks. Masi et .al. used a technique of perturbing the samples and creating multiple copies of images in 2D and in-plane augmentations in 3D  for face recognition \cite{masi2019face}. The VIHASI (Visual Human Action Silhouette) dataset for human action recognition was produced using virtual actors in a 3D environment to synthesize a variety of silhouette sequences. Data synthesis techniques like mirroring, scaling, translation, rotation, and noise addition are performed over the dataset to increase the diversity and variability of the dataset and thus handle the generalization challenge \cite{ragheb2008vihasi}. Data synthesis was performed by associating 2D images with 3D poses by selecting random image patches whose local 2D pose matched the projection of a given 3D pose and then by stitching them together kinematically \cite{rogez2016mocap}.  Mastorakis et.al used physics-based myoskeletal simulations to synthesize sequences of falls in adults by varying the height of the subject models \cite{mastorakis2018fall}. Physics-based musculoskeletal simulations were used to predict gait adaptations as a result of ankle plantar flexor muscle weakness and contractions \cite{Ong2019PredictingSimulations}.

% Gait based gender classification
%\subsection{Gait based gender classification}
%Gender is one of the essential cue in varied 
% applications. Identifying gait based gender unfolds numerous possibilities in varied sectors like security, surveillance, healthcare. Psychological experiments substatiate that humans by nature can identify gait based gender \cite{yu2009study}. Researchers perform gait based gender classifications from viewing angle variations \cite{upadhyay2022robust} Deep learning methods are pioneering in gait based gender classification approaches, that mostly depend on silhouettes and GEI's to classify male and female subjects\cite{lau2022tree} 

%Data Augmentation for Person Identification problem
%\subsection{Gait based Person Identification}
%Gait analysis aids in person reidentification.
%Being a non-invasive and subject collaboration-independent metric, it's widely applicable in numerous sectors. The gait analysis suffers from usual video or image-based challenges like occlusion, power image acquisitions, and most appropriate data. In the era of neural networks, human walking modeling is also accomplished using deep nets. Gaitset combines frames from varied scenarios \cite{chao2019gaitset}

Data augmentation methods have been used successfully on the image, time-series and video datasets. However, augmentation of gait datasets is challenging, as any newly generated data should respect the biomechanics of human walking. Prevailing methods synthesize human motion but the extent to which these synthetic motions depict near-natural human motion abiding by kinetic and kinematic constraints is highly debatable. We aim to address this issue by exploring physics-based biomechanical simulations to generate biomechanically plausible human motion.

\section{Biomechanical Data Augmentation} \label{sec:BiomechanicalDataAugmentation}

\begin{figure*}[!t]
\resizebox{13cm}{!}{
\centering
\includegraphics[height=8cm]{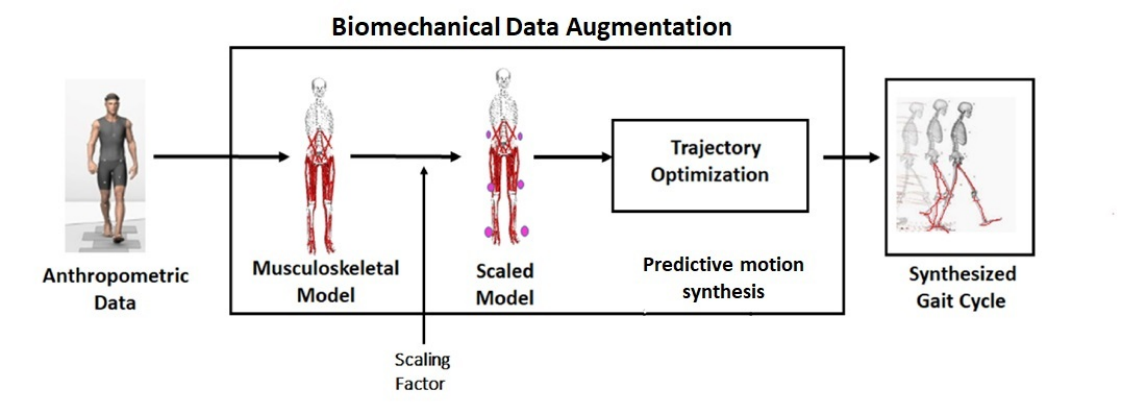}
\caption{Overview of the proposed framework for biomechanical data augmentation}
\label{Fig:BiomechanicalDataAugmentation}
}
\end{figure*}

\subsection{Musculoskeletal Modelling and Scaling} \label{MusculoskeletalModelling}
%Initially the scaling is performed over existing data to generate synthetic data.

Our approach aims to generate kinematically plausible gait data, in order to increase the size and variability of gait datasets. Data Augmentation is achieved using physics-based biomechanical simulation, based on anthropometric measurements. An overview of the proposed system for generating synthetic walking cycles is shown in Fig \ref{Fig:BiomechanicalDataAugmentation}. The subject's anthropometric measurements are fed as input to the model, which is scaled according to a range of factors. The scaled model is used to generate synthetic gait cycles abiding kinematical walking constraints. The following sections detail musculoskeletal modelling and scaling (section \ref{MusculoskeletalModelling}), followed by predictive motion synthesis (section \ref{Predictivemotionsynthesis}).

The locations of 3D joint positions on the lower extremity of a human subject, i.e. hip, knees, ankles and feet are the input to 
 our system. Such anthropometric data can be derived with the subject in either a standing pose, normally used for calibration or a walking pose as part of a gait capture sequence, using a motion capture system. % A pre-defined myoskeletal model \cite{ThelenOpenSimDocumentation}, used in OpenSim simulation software \cite{Delp2007OpenSim: Movement} representing human lower extremity anatomical attributes, joint frame locations, mass centre location, force application points, and muscle attachment points are chosen to ensure that its virtual markers can be associated to the joints of the human subject. The subject's lower extremity 3D joint positions are used to adapt the virtual markers of the chosen myoskeletal model to represent the considered subject's anthropometry.
 A pre-defined myoskeletal model \cite{ThelenOpenSimDocumentation} which is available in OpenSim \cite{Delp2007OpenSim:Movement}, an open-source software system for biomechanical modelling, simulation and analysis, has been utilised. This model represents various anatomical attributes of the human lower extremity, such as joint frame locations, mass centre locations, force application points, and muscle attachment points. The subject's lower extremity 3D joint positions are used to adapt the model to represent the subject’s anthropometry by associating its positions with the virtual markers of the model. 
The resultant modified model is further transformed using a range of scaling factors. These factors ensure that the body model is adjusted proportionally in all dimensions, resulting in a homogeneous scaling.

%Scaling Process
 The resultant modified model represents the specified subject's anthropometry and is further transformed using a range of scaling factors. Each scaling factor $s$ is applied uniformly and  homogeneously across all body segments. The model's geometry, joint frame locations, mass centre location, force application points, and muscle attachment points are all appropriately modified according to the scaling factor \cite{yu2020human}. 
 A lower extremity myoskeletal model $M$ in a static position is represented by its anthropometric measurements $A$. A scaled myoskeletal model $M_s$ is the aggregation of applying uniform scale factors $s$ across all anthropometric measurements.

\subsection{Predictive motion synthesis}\label{Predictivemotionsynthesis}

%reword the paragraph
Human walking is synthesized using physics-based biomechanical simulations. Specifically, gait is considered an optimal control problem, where muscle behaviour and consequently walking may be optimized according to goals such as human effort, joint loading, or locations of joints \cite{Dembia2019OpenSimControl}. Most popular datasets used specifically for gait-based recognition, like UPCV-gait \cite{kastaniotis2013gait}, CASIA-B \cite{yu2006framework}, OU-ISIR \cite{mori2010gait}, OUMVLP \cite{An_TBIOM_OUMVLP_Pose} lack information like human effort, muscle mass, kinetic information and others. In our approach, we perform predictive simulations of biological motion based on marker optimization over trajectory positions, %in turn, utilizing predefined joint loading and force distribution 
of myoskeletal models customized with only kinematic parameters and without kinetic parameters extracted from input video data. %and gait synthesis, joint loading needs force distribution, as our data is video-based, kinematic data is evident but not kinetic parameters
%We minimize the variation of the subject's center-of-mass speed from a target average speed.

Gait cycles are synthesized by trajectory optimization using either the direct collocation method (Moco) \cite{Dembia2019OpenSimControl} or the single shooting method (Scone) \cite{geijtenbeek2019scone}. 
The constraint here is to ensure the predicted motion is abided by normal human walking dynamics subject to gravity.
Moco implements the trapezoidal transcription of the direct collocation method and simulates motion for each specific anthropometric model, based on the target average speed and gait duration \cite{Dembia2019OpenSimControl}. %Scone performs trajectory optimization based on single shooting method that ensures optimizing parameters of a gait controller \cite{geijtenbeek2019scone}. 
Based on a predefined objective function the simulation software produces motion with a target to minimize metabolic cost, avoid falling and injury, and model movement with stabilized upper extremity \cite{Ong2019PredictingSimulations}.  
 
 %The trajectory optimization tool Scone is used for generating walking sequences \cite{DemsynthesizedSimControl}. 
 The Simulated Controller Optimization Environment (SCONE) \cite{geijtenbeek2019scone} is a trajectory optimization framework that uses the single shooting method to solve the  dynamic optimization problem for generating simulated data. It uses gait controllers to simulate the input myoskeletal model and optimize the parameters for the considered problem. We use Scone to simulate motion for each specific anthropometric model, based on the target's average speed and gait duration. The scaled OpenSim model is the input to Scone, and the gait controller performs gait simulation, in generating the next optimal position of the walking for the given duration.  % Write about walking simulation wrt to Scone

The scaled musculoskeletal model along with simulation parameters such as the target average speed and walking duration is supplied as input to the optimal solver \cite{wang2020impedance}. When the solver is initialized, the respective kinematic constraints in the musculoskeletal model are utilized. The holonomic scalar constraints of kinematics are added to the optimal control problem. Eventually, a sequence of human poses, each represented by a set of 3D joint coordinates and angles, for the given duration is generated. For each subject, multiple synthetic gait sequences may be derived by modifying the scale of their musculoskeletal model.

\section{Applications}\label{Applications}
The main purpose of augmenting gait datasets is to improve the performance of gait analysis based on machine learning by increasing the volume and variability of the training dataset. In order to explore the capabilities and limitations of the proposed data augmentation we consider two scenarios. Firstly,  3D skeleton gaits are augmented for the purpose of gender classification as a proof of concept of our approach (section \ref{Gait Classification}). Secondly, 2D skeleton gaits are augmented and used to train  person identification models (section \ref{PersonGaitIdentification}) to demonstrate the practical value of our approach to a popular gait-related application of computer vision.

\subsection{Gender Gait Classification}
\label{Gait Classification}

%\begin{figure*}[!t]
%\centering
%\includegraphics[width=14cm, height=7cm]%{GenderRecognitionPipeline.eps}
%\caption{\textcolor{red}{Gender Classification Pipeline}} 
%\label{Fig:Gender Classification}
%\end{figure*}

Human gender is one of the prominent features that play an evident role in individuals' gait patterns. We consider two classification approaches for gait-based gender classification, the Ensemble Subspace KNN classifier (ESKNN) \cite{Gul2018EnsembleClassifiers},  with superior performance among classical machine learning methods applied on human motion analysis problems \cite{ shehzad2023two}, and the LSTM classifier, a well-established deep learning method for analyzing time-sequences, which has been used by researchers for gait analysis tasks \cite{Kaushik2019EEG-BasedModel}.
%\cite{IEEESignalProcessingSociety2019France.}.

Every trial is represented by a vector $P$
\begin{equation}
P=\{p_{i,k}\}, i \in \{1,..,L\} , k \in \{1,..,K\} \label{eq:trial_vector}
\end{equation}

\noindent where  $p$ represents the respective joint angle, $i$ and $k$ are frame  and joint incices respectively, $L$ is the number of frames of the trial and $K$ is the number of joint angles used. The joint angles considered here are the hip flexion, knee flexion, and ankle flexion for each side of the body, so $K=6$. %add explanation for q and q in the next section

The feature vectors used in the ESKNN classifier are histogram descriptors similar to those in \cite{kastaniotis2013gait}. The histogram descriptor is aimed to efficiently encode the considered subject's gait information independently of the time taken or the number of frames and gait cycles. Therefore, normalized histograms for the K-selected lower extremity angles are built, each with  M bins:

\begin{equation}
H_k=[h_{\theta_1}, h_{\theta_2},...,h_{ heta_M}]\label{eq:histogram1}
\end{equation}
\noindent where $H_k \in \mathbb{R}^{M}, k \in \{1,..,K\}$.

Subsequently, all $K$ built histograms are normalized and concatenated to form a single feature vector per trial:

\begin{equation}
%H=[h\theta_1,h\theta_2...h\theta_{K-1},h\theta_{K}]
H=[\hat{H_1} \frown \hat{H_2} \frown ... \frown \hat{H_K}]\label{eq:histogram2}
 \end{equation}

\noindent where $H \in \mathbb{R}^{K \times M}$ summarises the gait.

The LSTM classifier is used with a five-layer structure comprising
a sequence layer, LSTM with hidden units, a fully connected layer, and softmax followed by the classification layer. The sequences are resampled to ensure a fixed length L for each trial for all feature vectors $P$ in the dataset. We also select the number of hidden units in the LSTM to be equal to $L$ to ensure sufficient memory for the whole walking trial. 

\subsection {Person Gait Identification} \label{PersonGaitIdentification}
\begin{figure*}[!t]
\resizebox{13cm}{!}{
\centering
\includegraphics[width=14cm, height=7cm]{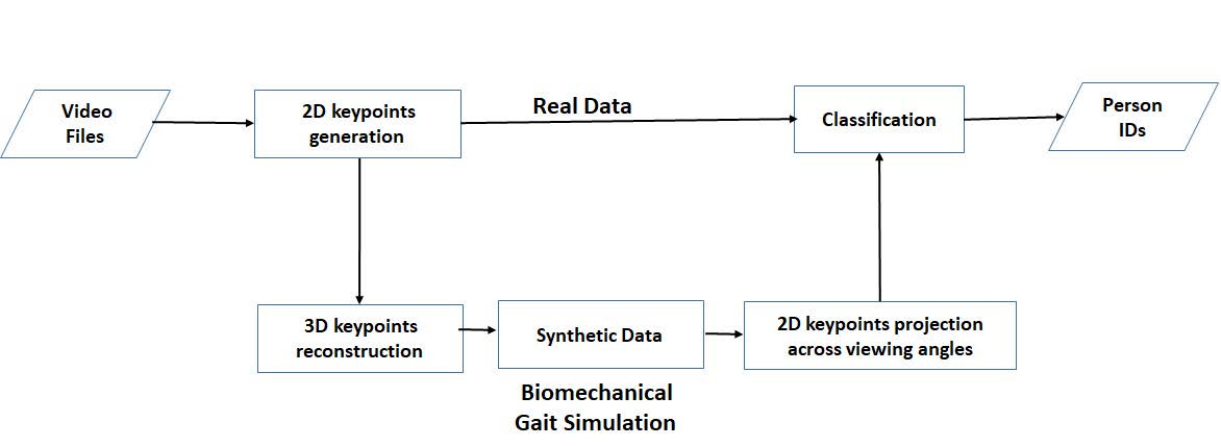}
\caption{Person identification pipeline} 
\label{Fig:PersonIdentification}}
\end{figure*}

To substantiate the value of biomechanical gait synthesis, we consider the application of person identification from gait sequences.
% Person identification pipeline briefing.
2D keypoints of the human lower extremity are extracted which form the real data for classification. A convolutional 3D human pose estimation method \cite{tome2017lifting} is used to reconstruct a sequence of 3D skeletons from a single sequence of 2D skeletons, which forms the input to our process of 3D biomechanical gait simulation as shown in Fig \ref{Fig:PersonIdentification}. The 3D data extracted using biomechanical data augmentation is reprojected on 2D using perspective projection to obtain feature vectors across varied viewing angles (as in  CASIA-B dataset) %and to compare results across varied viewing angles to the popular gait graph method \cite{teepe2022towards}. 
The feature vector obtained from real data along with the feature vector from synthetic data forms the basis for the person identification task. Here, we consider three classifiers to evaluate the efficiency of our method: LSTM, and Bi-LSTM which are popular deep learning methods for processing time series, human action recognition tasks \cite{dai2020human} and gait analysis  \cite{amin2021convolutional}, and the ResGCN network \cite{song2020stronger}, a method specific for skeleton-based motion analysis.

Every trial is represented using vector $Q$: 

\begin{equation}
Q=\{q_{i,k}\}, i \in \{1,..,L\} , k \in \{1,..,K\}
\label{eq:trial_vector2}
\end{equation}

 \noindent{where $q_{i,k}$ represent 2D coordinates, $i$ and $k$ are indices representing time and joint respectively, $L$ is the number of frames of the trial and $K$ the number of joint in 2D. The $K=6$ joints considered here are the hip, knee, and ankle for each side of the body. The feature vectors used here are the 2D joint coordinates for both the chosen classifiers.}

\section{Results} \label{Results} 
\begin{figure*}[!h]
\centering
\includegraphics[height=8cm]{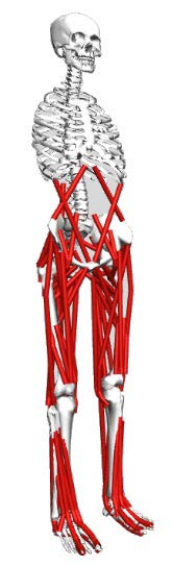}
\caption{Opensim model Gait2392 } 
\label{Fig:Gait2392}
\end{figure*}

We apply our data augmentation method on the publicly available datasets, Walking and Biomechanics Data Set (WBDS) (42 subjects) \cite{fukuchi2018public} and CASIA-B (124 subjects) \cite{yu2006framework}, for the tasks of gender classification and person identification respectively. We use the biomechanics simulation tool OpenSim \cite{delp2007opensim} to synthetically generate anthropometrically scaled musculoskeletal models using the pre-defined model Gait2392 \cite{ThelenOpenSimDocumentation} as shown in Fig \ref{Fig:Gait2392}, which closely associates to the considered datasets in combination with the two trajectory optimization tools, Moco \cite{Dembia2019OpenSimControl} and Scone \cite{geijtenbeek2019scone}.
For all experiments,  only real trials are used for testing, while real and/or simulated data may be used for training. Thus each subject's trial is considered for testing once and the metrics are calculated from the aggregated sum of values.

The anthropometry of each subject is derived from the captured static pose, used for calibration, and then consequently scaled using a range of scaling factors. The scale factor of 1.0 best fits the anthropometry of the specific subject to the musculoskeletal model. In our experiments, we consider seven scaling factors between (0.7 - 1.3) applied on the original subjects of WBDS and CASIA-B,  resulting in a total of 294 and 868 new synthesized anthropometric models respectively.

\subsection{Gender Classfication}
%WBDS dataset explanation

% we choose this dataset because..... rephrase
In order to evaluate the effectiveness of our methodology for gender classification based on gait, we choose the Walking Biomechanics dataset (WBDS). The WBDS consists of 3D motion capture (MoCap) data, encompassing demographics like age group, gender, walking speed, and limited subject information, alongside additional metadata. Through the utilization of the WBDS, our objective was to assess the accuracy of our approach in accurately categorizing gender based on gait patterns. The dataset is further a comprehensive set, including kinetic and kinematic variables, measured during both overground and treadmill walking with high-quality data. The  inclusion of relevant biomechanical variables such as joint positions and speed variations, as well as a smaller number of subjects, makes the dataset well-suited for our specific objective of biomechanical gait simulation. For this evaluation, we considered the overground walking trials of the subjects from the dataset. This dataset contains barefoot walking trials of 42 healthy individuals, including 24 males (height 172.9 $\pm$ 8.4cm, mass 73.68 $\pm$ 8.2 kg)  and 18 females (height 158.6 $\pm$ 8.1cm, mass 59.88 $\pm$ 9.6 kg ), performed on a flat surface at a range of gait velocities (1.3 $\pm$ 0.25 m/s). The subjects can also be split according to their age group, i.e. 24 young adults (age 27.6 $\pm$ 4.4 years, height 171.1 $\pm$ 10.3cm and mass 68.4 $\pm$ 11.7kg) and 18 older adults (age 62.7 $\pm$ 8 years, height 161.8 $\pm$9.2cm and mass 66.9 $\pm$9.78kg). All participants performed overground walking trials at self-selected comfortable speeds, then at approximately  30\% faster and 30\% slower speeds. Several trials for each speed category and for each participant were recorded along with their gait velocities resulting in 1641 trials overall. A total of 1641 trial velocities of various subjects (1.3 $\pm$ 0.25 m/s), taken from the WBDS dataset, was provided as the input to synthesize multiple walking trials, with a uniform walking duration of 2 seconds per each trial. Considering the seven anthropometrically scaled musculoskeletal models (0.7-1.3) %for varied target speed values from the dataset (with an average of 0.8 m/s)
for each subject as mentioned in the dataset, a total of 11,487 trials were synthesized.

We use the methods for gait-based gender classification mentioned in section \ref{Gait Classification} to check the validity of synthesized gaits. We use M=20 bins per angle for the histogram presentation and L=100 frame fixed length for the sequence representation to formulate the input feature vectors used by the ESKNN and the LSTM classifiers respectively. The classification is performed by adopting the leave-one-out subject protocol, i.e. for each experiment we reserve the trials of a particular subject for testing and use the rest for training.

\begin{comment}\begin{figure*}[!h]
\centering
\includegraphics{FigGenderClassification.png}
\caption{Gender Classification } 
\label{Fig:GenderClassificationLineChart}
\end{figure*}
\end{comment}

\begin{comment}
\begin{figure*}[!h]
\centering
\includegraphics{FigBarChart4_GenderClassification.png}
\caption{Gender classification accuracies comparison with real and simulated data from Moco and Scone } 
\label{Fig:GenderClassificationBarChart}
\end{figure*}
\end{comment}
% Gender Classification results
\begin{table}[!h]
\renewcommand{\arraystretch}{0.8}
\caption{\textbf{Gender classification weighted accuracy percentage}}
\label{table2}
\centering
%\begin{tabular}{|p{3cm}|p{2.5cm}|p{2.5cm}|p{2.5cm}|}
\begin{tabular}{|c|c|c|c|}
\hline
\multicolumn{1}{|m{4cm}|}{\textbf{Training Data}} & \textbf{Testing} & \multicolumn{1}{|m{1.5cm}|}{\textbf{ESKNN}} & {\textbf{ LSTM}}\\
\hline
\multicolumn{1}{|m{4cm}|}{Real} & Real & 79.30 & 86.00\\
\hline
\multicolumn{1}{|m{4cm}|}{Simulated (Moco)}& {Real} & {81.30} & {81.00}\\
\hline
\multicolumn{1}{|m{4cm}|}{Real + Simulated (Moco)} & {Real} & {82.10}&{92.30} \\
\hline
\multicolumn{1}{|m{4cm}|}{Simulated (Scone)} & Real &86.30 & 92.80\\
\hline
\multicolumn{1}{|m{4cm}|}{Real + Simulated (Scone)} & Real & \textbf{87.50} &\textbf{\underline{94.60}} \\
\hline
\end{tabular}
\end{table}

The performance of gender classification using either Moco or Scone for trajectory optimization is given in Table \ref{table2} and Table \ref{GenderClassificationtable2}. The results show that the classification accuracy improves when including the simulated data in the training. Using Scone for trajectory optimization leads to more representative gait sequences than using Moco. Interestingly, training using only simulated data generated by Scone leads to better performance than training with real data in most cases, which can be justified by the amount and variability of the simulated data, i.e. seven times the amount of real data. Table \ref{GenderClassificationtable2} results affirm our claim. From the results, we could infer that LSTM performs better for gender classification task, with weighted accuracies ranging from 86\% to 94.6\% when training with real-only and with a combination of real and synthetic data.  For all the remaining experiments, we will use the Scone method for trajectory optimisation in our biomechanics-based augmentation.

In all cases, combining real and synthesized trials in the training dataset improves the performance of gait classification. The cumulative trials generated with Scone for classification shown in Table \ref{GenderClassificationtable2} justifies this. Results also confirmed that the more synthetic data the better improvement in accuracy. %Even in the case of age classification where LSTM, trained only with real trials achieves 96.0\% and the room for improvement is fairly small, adding simulated trials produced by Scone leads to further improvement. 

%Table with cumulative results for gait classification
\begin{table}[!h]
\renewcommand{\arraystretch}{0.8}
\caption{\textbf{Gender classification using Scone with various scale factors. The best result for each method is highlighted in bold and the overall best result is underlined.}}
\label{GenderClassificationtable2}
\centering
%\begin{tabular}{|p{3cm}|p{2.5cm}|p{2.5cm}|p{2.5cm}|}
\begin{tabular}{|c|c|c|c|}
\hline
\multicolumn{1}{|m{4.5cm}|}{\textbf{Training Data}} & \textbf{Testing} & \multicolumn{1}{|m{1.5cm}|}{\textbf{ESKNN}} & {\textbf{ LSTM}}\\
\hline
\multicolumn{1}{|m{4.5cm}|}{Real only} & Real & 79.30 & 86.00 \\
\hline
\multicolumn{1}{|m{4.5cm}|}{Real + Simulated (s=1.0)} & Real & 87.16 & 93.33\\
\hline
\multicolumn{1}{|m{4.5cm}|}{Real + Simulated (s=0.9:1.1)} & Real &87.23 & 93.98 \\
\hline
\multicolumn{1}{|m{4.5cm}|}{Real + Simulated (s=0.8:1.2)} & Real & 87.48 & 94.06 \\
\hline
\multicolumn{1}{|m{4.5cm}|}{Real + Simulated (s=0.7:1.3)} & Real & \textbf{87.50}& \textbf{\underline{94.60}} \\
\hline
\end{tabular}
\end{table}

\subsection{Person Identification}
% Casia B datasets explanation
CASIA-B \cite{yu2006framework} dataset that has been predominantly used to evaluate gait-related tasks in the literature is chosen for our person identification experiments. The dataset consists of walking sequences of 124 individuals under 3 walking conditions: normal walking (NM) sequences, walking with bag (BG), and walking with a jacket or a coat (CL)  across 11 different viewing angles. Each subject performs 6 normal walking (NM) sequences, 2 sequences of walking with a bag (BG), and 2 sequences of walking with wearing a jacket or a coat (CL). The normal sequences in RGB are used to extract the subject's anthropometric information and then to generate synthetic gait, abiding by biomechanical constraints. Therefore, a total of 66 sequences are utilized as the base for the person identification task.

\begin{figure*}[!h]
\centering
\includegraphics[height=8cm]{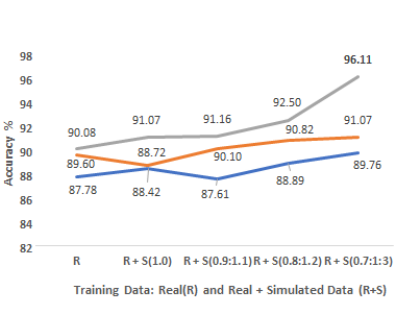}
\caption{Gait person identification accuracy for different sets of data augmentation} 
\label{Fig:PersonIdentificationLineChart}
\end{figure*}

% Moved to the beginning of this section
%We use the modelling and movement simulation tool OpenSim \cite{delp2007opensim} to synthetically generate anthropometrically scaled musculoskeletal models using the pre-defined model Gait2392 \cite{ThelenOpenSimDocumentation} as shown in Fig \ref{Fig:Gait2392}, which closely associates to the considered datasets in combination with one of the two trajectory optimization tools, \textcolor{green}{\textbf{Moco}} and Scone.

%The anthropometry of each subject is derived from the captured static pose, used for calibration, and then consequently scaled using a range of scaling factors. The scale factor of 1.0 best fits the anthropometry of the specific subject to the musculoskeletal model. In our experiments, we consider seven scaling factors (0.7-1.3) resulting in a total of 868 new synthesized anthropometric models CASIA-B dataset.
%With WBDS dataset a total of 1641 trial velocities of various subjects (1.3 $\pm$ 0.25 m/s), taken from the WBDS dataset, was provided as the input to synthesize multiple walking trials, with a uniform walking duration of 2 seconds per each trial. Considering the six anthropometrically scaled musculoskeletal models for a range of target speeds for each subject, a total of 9,846 trials were synthesized.

%casia B
The inputs for generating  multiple synthesized walking trials using Scone were 868 (124 subjects x 7 scale factors) synthesized anthropometric models %were the inputs to synthesize multiple walking trials using Scone for trajectory optimization,
with a uniform walking duration of 2 seconds in each trial resulting in a total of 9,548 (868 models x 11 views) synthesized trials. 

The results of gait-based person identification as shown in Fig \ref{Fig:PersonIdentificationLineChart}, confirm that augmenting the training set with the biomechanically synthetic data leads to increased performance classification accuracy for all three algorithms. The more the simulated sequences are added to the training set, the more accuracy increases. As shown in Table \ref{PersonIdentification}, the training with simulated data only %with seven scaled trials 
in the training is %mostly
on par with the training with real data only with an accuracy of 87.21\% and 90.34\% for LSTM and Bi-LSTM classifiers against real data with 88.72\% and 90.88\% over both machine learning classifiers. When combining real and simulated data, the accuracy increases distinctively, validating the presence of synthetic data, and elevating the accuracy to 91.07\% and 96.11\% with former and latter classifiers respectively. On comparing our results to the ResGCN \cite{song2020stronger} method, which has been used in GaitGraph \cite{teepe2021gaitgraph} using the CASIA-B dataset,  the outcome clearly indicates that by adding biomechanically simulated gait data, generated using our method, significantly improves the accuracy from 87.6\% to 89.77\% respectively.

% Include an explanation about the diagram real vs simulated
 % The simulated trials of a male and a female subject for a specific gait speed for the four scaling factors are shown in Fig.\ref{Fig:FourScaleFactorsMale} and Fig.\ref{Fig:FourScaleFactorsFemale} respectively, along with the real gait. The simulated key gait parameters (knee and hip flexion angles) seem to follow the kinematic patterns of normal human motion as defined by Rose et al. \cite{rose1994human}.

%addidng Person identification table
\begin{table}[!h]
\renewcommand{\arraystretch}{0.7}
\caption{\textbf{Person identification accuracy with best results highlighted. All simulated data has arrived from Scone and the best results are highlighted}}
\label{PersonIdentification}
\centering 
%\begin{tabular}{|p{3cm}|p{2.5cm}|p{2.5cm}|p{2.5cm}|}
\begin{tabular}{|c|c|c|c|c|}
\hline
\multicolumn{1}{|m{3.5cm}|}{     \textbf{Training Data}} & \textbf{Testing} & \multicolumn{1}{|m{1.5cm}|}{\textbf{LSTM}} & {\textbf{ Bi-LSTM}} & {\textbf{ ResGCN}}\\
\hline
\multicolumn{1}{|m{4.0cm}|}{Real} & Real &  88.72& 90.88 &87.78\\

\hline
\multicolumn{1}{|m{4.0cm}|}{Simulated (Scone)} & Real & 87.21 & 90.34 & 88.77\\
\hline
\multicolumn{1}{|m{4.0cm}|}{Real +  Simulated (Scone)} & Real & \textbf{91.07} & \textbf{96.11} &\textbf{89.76}\\
\hline
\end{tabular}
\end{table}

Table \ref{PersonidentificationusingBi-LSTMacrossviews} shows the comparison of person identification results  across 11 viewing angles and the averaged mean score for the three classifiers considered. The person identification accuracy in most cases is comparatively higher when trained with real and simulated data compared to real only, confirming the value of our data augmentation approach. %Notable, Bi-LSTM showing higher accuracy than others. %Classification scores using real and simulated data using our method outperforms the state-of-art ResGCN  method, across all 11 views and the mean score.
%two widely used algorithms LSTM and BI-LSTM. Also, the table showcases the comparison of our method to ResGCN, another proven method for person identification using the same dataset, CASIA-B. The results clearly indicate that our method outperforms ResGCN and also results in increased accuracies across all 11 cross-views. 

\begin{figure*}[!t]
\centering
\includegraphics[height=8cm]{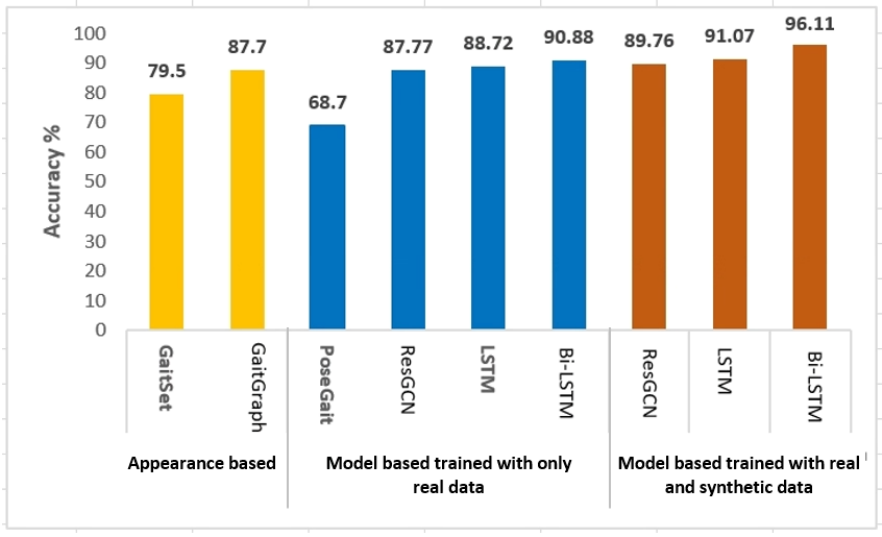}
\caption{Accuracies in percent on CASIA-B comparing our method, Biomechanical Data Augmentation(BDA) with appearance-based and model-based methods} 
\label{Accuraciesinpercent}
\end{figure*}

%The best-performing models in the literature Gaitset \cite{chao2019gaitset} and Gaitnet \cite{song2019gaitnet} are appearance-based methods.
We compare data-augmented models, with appearance-based and other model-based methods (Fig \ref{Accuraciesinpercent}). As discussed before, all data-augmented versions of model-based methods outperform the original versions, including GaitGraph. In addition, our data augmentation delivers state-of-the-art results, even against appearance-based methods. The results clearly indicate that simulated data using our biomechanical gait augmentation method along with the real data assist model-based person identification methods to achieve superior results on the CASIA-B dataset. %Furthermore, data-augmented version of  ResGCN network outperforms GaitGraph. %The outcome of experiments further emphasises that model-based methods outperform appearance-based methods.

\begin{flushleft}
\begin{table}[!h]
\renewcommand{\arraystretch}{1.0}
\resizebox{15cm}{!} {
\caption{\textbf{Accuracy in percent across 11 viewing angles and average ranked accuracy on Casia-B for LSTM, Bi-LSTM and ResGCN. The data column represents the data set used for training, where R represents real data only and S represents real augmented with simulated data. The best results are highlighted}}
\label{PersonidentificationusingBi-LSTMacrossviews}
\centering
\begin{tabular}{|p{1.3cm}|p{0.6cm}|p{0.7cm}|p{0.7cm}|p{0.7cm}|p{0.7cm}|p{0.7cm}|p{0.7cm}|p{0.7cm}|p{0.7cm}|p{0.7cm}|p{0.7cm}|p{0.7cm}|p{0.9cm}|}
\hline

%&& & & & & & & & & & & \\
\textbf{\small{Method}} & \textbf{\small{Data}} & {\textbf{0}} & {\textbf{18}} & {\textbf{36}}& {\textbf{54}}& {\textbf{72}}& {\textbf{90}}& {\textbf{108}}& {\textbf{126}}& {\textbf{144}}& {\textbf{162}}& {\textbf{180}}& 
\textbf{\small{Mean}}
\\[0.2cm]
%&& & & & & & & & & & & \\
\hline
%&& & & & & & & & & & & \\
\multirow{2.5}{*}{\textbf{\small{LSTM}}} & \textbf{R} &89.33& 87.97&88.67& 87.92&88.86&90.89&88.66&87.56&88.92&87.52&89.57 & 88.72
\\ [0.2cm]
%&& & & & & & & & & & & \\ 

\cline{2-14}
%&& & & & & & & & & & & \\ 
\textbf{} & \textbf{S} &\textbf{91.17}&\textbf{91.70}&\textbf{91.27}&\textbf{89.55}&\textbf{90.98}&\textbf{91.45}&\textbf{90.01}&\textbf{91.45}&\textbf{90.05}&\textbf{92.01}&\textbf{91.57}&\textbf{91.07}
\\[0.2cm]
%&& & & & & & & & & & & \\
\hline

%&& & & & & & & & & & & \\
\multirow{2}{*}{\textbf{\small{Bi-LSTM}}} & \textbf{R} &  89.94 & 88.22 & \textbf{\underline{94.80}} & 94.40 & 90.93 & 95.42& 89.22& 83.93 & 89.19 & 90.28 & 93.39 & 90.88
\\[0.2 cm]
%&& & & & & & & & & & & \\
\cline{2-14}
%&& & & & & & & & & & & \\ 
\textbf{} & \textbf{S} & \textbf{\underline{94.36}}& \textbf{\underline{95.11}}&94.38&\textbf{\underline{96.79}}&\textbf{\underline{94.67}}&\textbf{\underline{98.89}}&\textbf{\underline{97.02}}&\textbf{\underline{95.67}}&\textbf{\underline{97.21}}&\textbf{\underline{95.46}}&\textbf{\underline{97.68}}&\textbf{\underline{96.11}}
\\[0.2 cm]
%&& & & & & & & & & & & \\
\hline
%\hline
%&& & & & & & & & & & & \\
\multirow{2}{*}{\textbf{\small{ResGCN}}} & \textbf{R} &87.15&87.62&86.96&87.11&87.43&86.92&89.77&86.72&\textbf{87.90}&87.34&88.79&87.60\\ [0.2 cm]
%&& & & & & & & & & & & \\ 
\cline{2-14}
%&& & & & & & & & & & & \\ 
\textbf{} & \textbf{S} & \textbf{89.10}&\textbf{90.15}&\textbf{89.02}&\textbf{89.35}&\textbf{88.40}&\textbf{91.30}&\textbf{90.75}&\textbf{90.90}&86.43&\textbf{89.35}&\textbf{92.69}&\textbf{89.77}\\[0.2cm]
%&& & & & & & & & & & &\\
\hline
\end{tabular}}
\end{table}
 \end{flushleft}

\begin{comment}
\begin{table}[!h]
\renewcommand{\arraystretch}{0.8}
\caption{\textbf{Accuracies on CASIA-B across various methods for person identification}}
\label{table7}
\centering
\small
%\begin{tabular}{|p{3cm}|p{2.5cm}|p{2.5cm}|p{2.5cm}|}
\begin{tabular}{|c|c|c|}
\hline
\multicolumn{1}{|m{2cm}|}{\textbf{Type}}& \multicolumn{1}{|m{2cm}|}{\textbf{Method}} & \textbf{Accuracy}\\
\hline 
\multicolumn{1}{|m{2cm}|}{Appearance}& \multicolumn{1}{|m{2cm}|}{GaitSet\cite{chao2019gaitset}}  & 95\\ %\cite{chao2019gaitset}
\multicolumn{1}{|m{2cm}|}{-based}& \multicolumn{1}{|m{2cm}|}{Gaitnet \cite{song2019gaitnet}} & 91.6\\ %\cite{song2019gaitnet} 
\hline
&  &  \\
 &\multicolumn{1}{|m{2cm}|}{Gaitgraph} & 87.7\\ %\cite{teepe2021gaitgraph} 
 \multicolumn{1}{|m{2cm}|}{Model based} & 
 \multicolumn{1}{|m{2cm}|}{} & \\ 
\multicolumn{1}{|m{2cm}|}{}&\multicolumn{1}{|m{2cm}|}{PoseGait} & 68.7 \\
 &  &  \\

\hline
 &  & \\
\multicolumn{1}{|m{2cm}|}{} & \multicolumn{1}{|m{2cm}|}{ResGCN} & \textbf{89.76} \\
\multicolumn{1}{|m{2cm}|}{Our Methods} & \multicolumn{1}{|m{2cm}|}{LSTM} & \textbf{91.07} \\
\multicolumn{1}{|m{2cm}|}{} & \multicolumn{1}{|m{2cm}|}{Bi-LSTM} & \textbf{96.11} \\

 &  & \\
\hline
\end{tabular}
\end{table}
\end{comment}

\section{Conclusion} \label{conclusion}
This paper proposed a novel framework for gait data augmentation by using physics-based simulations to synthesize biomechanically plausible walking sequences, aiming to address the issue of scarcity of 3D and 2D gait data. Specifically, the anthropometric features were slightly varied to augment existing gait using OpenSIM and Scone. 

Our approach was validated by applying the proposed data augmentation framework on the 3D MoCap WBDS dataset and the 2D CASIA-B multi-view video dataset for gait-based gender classification and person identification for a range of classifiers such as ESKNN, LSTM, Bi-LSTM and ResGCN. Experimental results demonstrated a clear improvement in accuracy when the data was augmented for all gait-based classifiers tested. Our version of Bi-LSTM person identification model, trained with our augmented version of CASIA-B achieved state-of-the-art results of 96.11\%. 

A possible extension of this work is augmenting data expressed in other modalities and features like depth, silhouette, and kinetic such as ground reaction forces and contact forces along with the kinematic data. Another area for future work is the synthesis of physically impaired gaits.

\section{Acknowledgement}
The authors would like to thank Kingston University for partially supporting this research through a Doctoral Training Alliance (DTA) PhD Studentship.

%%%%%%%%% REFERENCES

\bibliographystyle{apalike}
\bibliography{references.bib} 
\end{document}